\documentclass[lettersize,journal]{IEEEtran}
\usepackage{amsmath,amsfonts}
\usepackage{algorithmic}
\usepackage{algorithm}
\usepackage{array}
\usepackage[caption=false,font=normalsize,labelfont=sf,textfont=sf]{subfig}
\usepackage{textcomp}
\usepackage{stfloats}
\usepackage{url}
\usepackage{verbatim}
\usepackage{graphicx}
\usepackage{cite}
\begin{document}

\title{Funabot-Upper: McKibben Actuated Haptic Suit Inducing Kinesthetic Perceptions in Trunk, Shoulder, Elbow, and Wrist}

\author{Haru Fukatsu$^{1}$, Ryoji Yasuda$^{2}$, Yuki Funabora$^{1}$, and Shinji Doki$^{1}$
        % <-this % stops a space
\thanks{*This work was supported by the JST Japan Primary Research Support Program (JPMJFR216T) and by the Young Scientists Initiative Research Unit of Nagoya University.}% <-this % stops a space
\thanks{$^{1}$ Department of Information and Communication Engineering, Graduate School of Engineering, Nagoya University, Fro-cho, Chikusa-ku, Nagoya, Aichi, Japan}
\thanks{$^{2}$ Department of Electrical Engineering, Electronics, and Information Engineering, School of Engineering, Nagoya University, Fro-cho, Chikusa-ku, Nagoya, Aichi, Japan}}

% The paper headers
%\markboth{Journal of \LaTeX\ Class Files,~Vol.~14, No.~8, August~2021}%
%{Shell \MakeLowercase{\textit{et al.}}: A Sample Article Using IEEEtran.cls for IEEE Journals}

%\IEEEpubid{0000--0000/00\$00.00~\copyright~2021 IEEE}
% Remember, if you use this you must call \IEEEpubidadjcol in the second
% column for its text to clear the IEEEpubid mark.

\maketitle

\begin{abstract}
This paper presents Funabot-Upper, a wearable haptic suit that enables users to perceive 14 upper-body motions, including those of the trunk, shoulder, elbow, and wrist. Inducing kinesthetic perception through wearable haptic devices has attracted attention, and various devices have been developed in the past.
However, these have been limited to verifications on single body parts, and few have applied the same method to multiple body parts as well.
In our previous study, we developed a technology that uses the contraction of artificial muscles to deform clothing in three dimensions.
Using this technology, we developed a haptic suit that induces kinesthetic perception of 7 motions in multiple upper body.
However, perceptual mixing caused by stimulating multiple human muscles has occurred between the shoulder and the elbow.
In this paper, we established a new, simplified design policy and developed a novel haptic suit that induces kinesthetic perceptions in the trunk, shoulder, elbow, and wrist by stimulating joints and muscles independently.
We experimentally demonstrated the induced kinesthetic perception and examined the relationship between stimulation and perceived kinesthetic perception under the new design policy. Experiments confirmed that Funabot-Upper successfully induces kinesthetic perception across multiple joints while reducing perceptual mixing observed in previous designs. The new suit improved recognition accuracy from 68.8\% to 94.6\% compared to the previous Funabot-Suit, demonstrating its superiority and potential for future haptic applications.

\end{abstract}


\section{Introduction}
\IEEEPARstart{I}{n} recent years, wearable haptic devices that can induce kinesthetic perception have attracted attention. 
For example, Teranishi et al. created device for shoulder using kinesthetic illusions caused by vibratory stimulation of muscle tendons\cite{Vibe3}. Rangwani et al. conducted research on a device for elbow motion using transcutaneous electrical stimulation\cite{rangwani2021new}. Nakamura et al. developed haptic device with shear force to upper limb using the hanger reflex\cite{Skin2}.
Similarly, many studies have developed wearable haptic devices for upper body.
There are a lot of methods to induce kinesthetic perception, such as vibration\cite{Wrist1_Vibe1,Vibe2,Vibe3,Vibe4}, skin deformation\cite{Skin1,Skin2,Skin3,Skin4}, and electrical stimulation\cite{electrical1,rangwani2021new}.
In addition, some wearable haptic devices have been developed in various fields such as the trunk\cite{Trunk1,Trunk2}, shoulder\cite{Shoulder1,Shoulder2}, elbow\cite{Elbow1,Elbow2}, and wrist\cite{Wrist1_Vibe1,Wrist2}.
However, most devices focus on specific motions of single body part. In addition, while they are good at transmitting accurate motions or providing sensations in those limited body parts, there are few studies that have verified whether it is possible to induce kinesthetic perception in multiple parts of the upper body using the same device.
One of the reasons for this may be design-related factors, such as device rigidity and weight, the need for tuning according to the subject, or complicated systems.

We have developed a suit type system that uses the contraction of artificial muscles to deform clothing in three dimensions, providing intuitive kinesthetic perception by stimulating the wearer's skin\cite{Funabot1,Funabot2}. Using this technology, we have developed many kinds of wearable haptic suits and conducted experiments to verify whether the same method can be adaptive in multiple parts of the upper body.

First, we studied a suit that can induce kinesthetic perception for the four motions of the trunk\cite{Suit}. 
Second, we studied a suit for inducing elbow flexion and extension\cite{yokoe2024elbow}, and further studied a suit capable of inducing six motions in total by adding four shoulder motions\cite{yokoe2025intuitive}.
Finally, we previously developed Funabot-Suit for Upper Body, a wearable haptic suit that can induce kinesthetic perception of upper body from global body part such as the trunk to local body part such as the shoulder and the elbow using a single haptic suit\cite{Upper}.

Funabot-Suit for Upper Body was designed as a suit capable of inducing kinesthetic perception in 7 motions of the trunk, shoulder, and elbow. 
Through experiments with subjects, we verified whether stimulation of each part was not mixed with stimulation of other parts and confirmed that the trunk could be identified without any mixing with other parts.
However, the results of the experiment on the shoulder and elbow were not as favorable as expected and there was a problem of perceptual mixing between those areas.
This perceptual mixing in the Funabot-Suit for Upper Body was likely caused by its design policy. The arrangement of artificial muscles in previous studies was determined through trial and error to mimic human muscles.

In this study, we developed a novel haptic suit: Funabot-Upper. It utilizes the same technology as previous haptic suits but has different design policy. Funabot-Upper can induce 7 motions of previous Funabot-Suit (trunk left/right rotation, shoulder adduction/abduction/horizontal adduction, and elbow flexion/extension) and new 7 motions (trunk forward/backward bending, shoulder adduction, and wrist adduction/abduction/flexion/extension), total of 14 motions with a single suit.
In this paper, we proposed new design policy and developed Funabot-Upper based on it.
We confirmed whether Funabot-Upper can induce kinesthetic perception and reduce perceptual mixing by conducting a subject experiment. In addition, we compared and evaluated the performance between Funabot-Upper and Funabot-Suit for Upper Body.

There are three contributions of this study.
First, we established a new, simplified design policy and developed a novel haptic suit that targets 14 motions of entire upper body. Through subject experiment using this suit, we were able to confirm the validity of the hypothesis of perceptual mixing.
Second, we experimentally clarified the effects obtained by stimulating joints and muscles independently.
Third, we experimentally clarified that Funabot-Upper can induce kinesthetic perception to wearers more accurately than Funabot-Suit for Upper Body.
These findings will contribute to design guidelines for future haptic devices.

\section{Method}
In this section, we introduce the new design policy consisting of two factors (separation of stimulation, and system simplification), and Funabot-Upper based on it.

\begin{figure*}[!t]
    \centering
    \vspace{2mm}
    \includegraphics[width=1\linewidth]{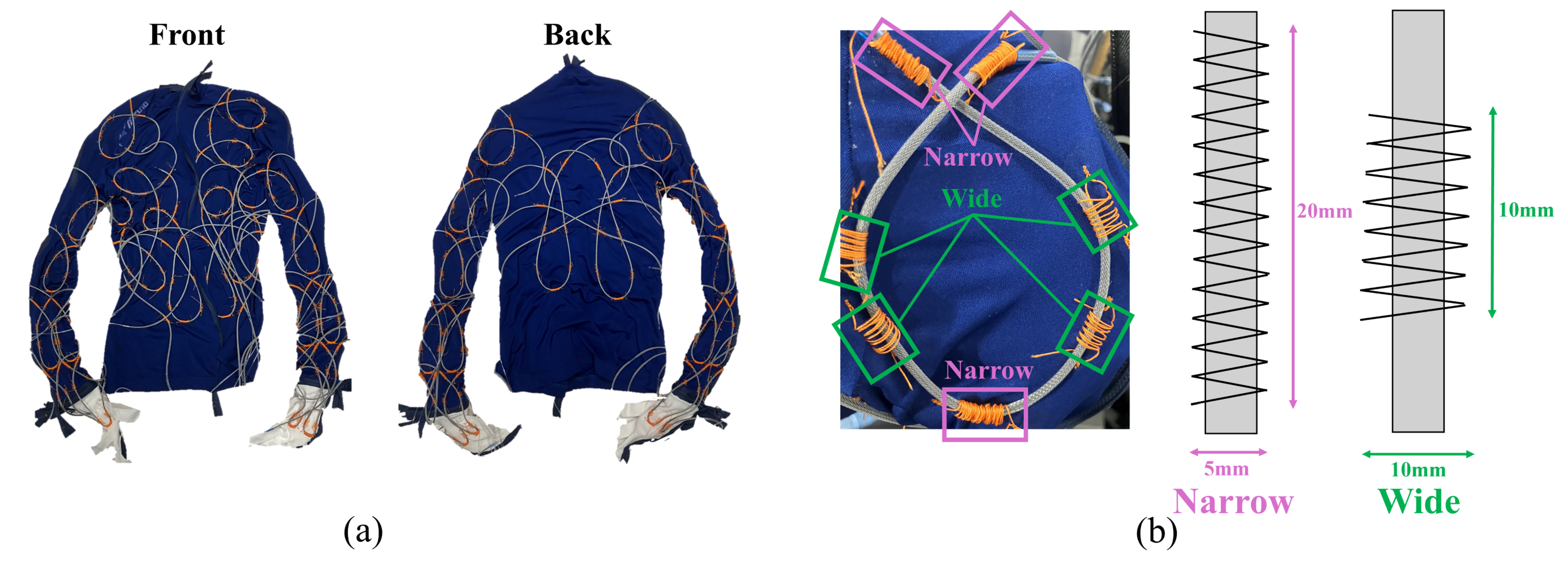}
    \caption{(a) Appearance of Funabot-Upper. Underwear and gloves are connected through artificial muscles. (b) Two different embroidery methods depending on the embroidery width to secure the artificial muscle to the clothing surface.}
    \label{g1}
\end{figure*}

\subsection{The Design Policy of Funabot-Upper}
\begin{figure*}[!t]
    \centering
    \vspace{2mm}
    \includegraphics[width=0.9\linewidth]{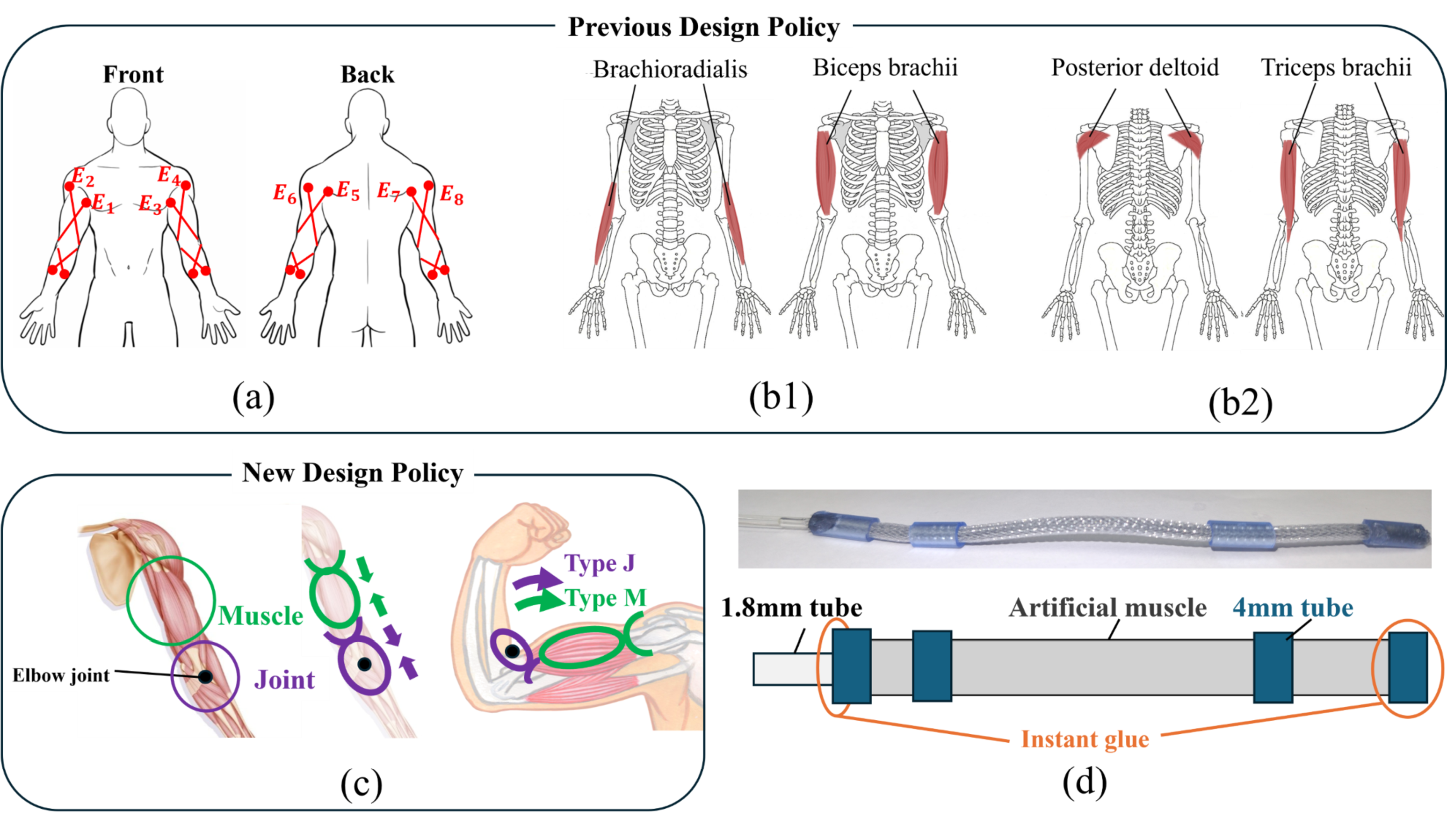}
    \caption{(a)The arrangement of artificial muscles for inducing elbow flexion in Funabot-Suit for Upper Body\cite{Upper}. (b1)The human muscles mimicked by the arrangement of artificial muscles in Fig. \ref{design}(a)\cite{Upper}. (b2)The human muscles stimulated other than those in Fig. \ref{design}(b1) (Posterior Deltoid contribute to shoulder horizontal extension and Triceps Brachii contribute to elbow extension)\cite{Upper}. (c)New design policy: Stimulate the joints and muscles independently. (d)Artificial muscle configuration. It is sealed the air escape route at the endpoints using instant glue.}
    \label{design}
\end{figure*}

The perceptual mixing observed in the previous Funabot-Suit was likely caused by artificial muscles stimulating multiple human muscles simultaneously.
For example, when inducing elbow flexion, artificial muscles were arranged along the upper arm and forearm and activated brachioradialis and biceps brachii as shown in Fig. \ref{design}(a)(b1).
However, it also activated the surrounding muscles such as the posterior deltoid and triceps brachii as shown in Fig. \ref{design}(b2), resulting in ambiguous sensations. 
To reduce such interference, Funabot-Upper separates joint stimulation from muscle stimulation.

In addition, we tried to simplify the fabrication of suit. Funabot-Suit for Upper Body utilized 80 artificial muscles in order to induce 7 motions.
When increasing the number of induced motions, this number of artificial muscles may complicate the system. Thus, we also aim to reduce the number of artificial muscles.

Based on the above, the design policy for Funabot-Upper emphasizes independent stimulation of joints and muscles to minimize perceptual interference.

Each stimulus is classified as either:
\begin{itemize}
    \item Joint stimulus (Type J): positioned across a human joint and applying force along the joint's motion direction.
    \item Muscle stimulus (Type M): positioned along a human muscle and applying shear or compression force parallel to the muscle.
\end{itemize}

By separating these two categories, we aim to isolate joint rotation stimulus from muscle tension stimulus.

Each joint or muscle can be basically assigned up to one artificial muscle for a single motion.
The arrangement shape of artificial muscle was modified to achieve the effect of its contraction while limiting the stimulation range. The arrangement of artificial muscles was basically designed in an elliptical shape. The effect of that shape will be mentioned in section II.B.

\subsection{The Configuration of Funabot-Upper}
\begin{figure*}[!t]
    \centering
    \includegraphics[width=0.7\linewidth]{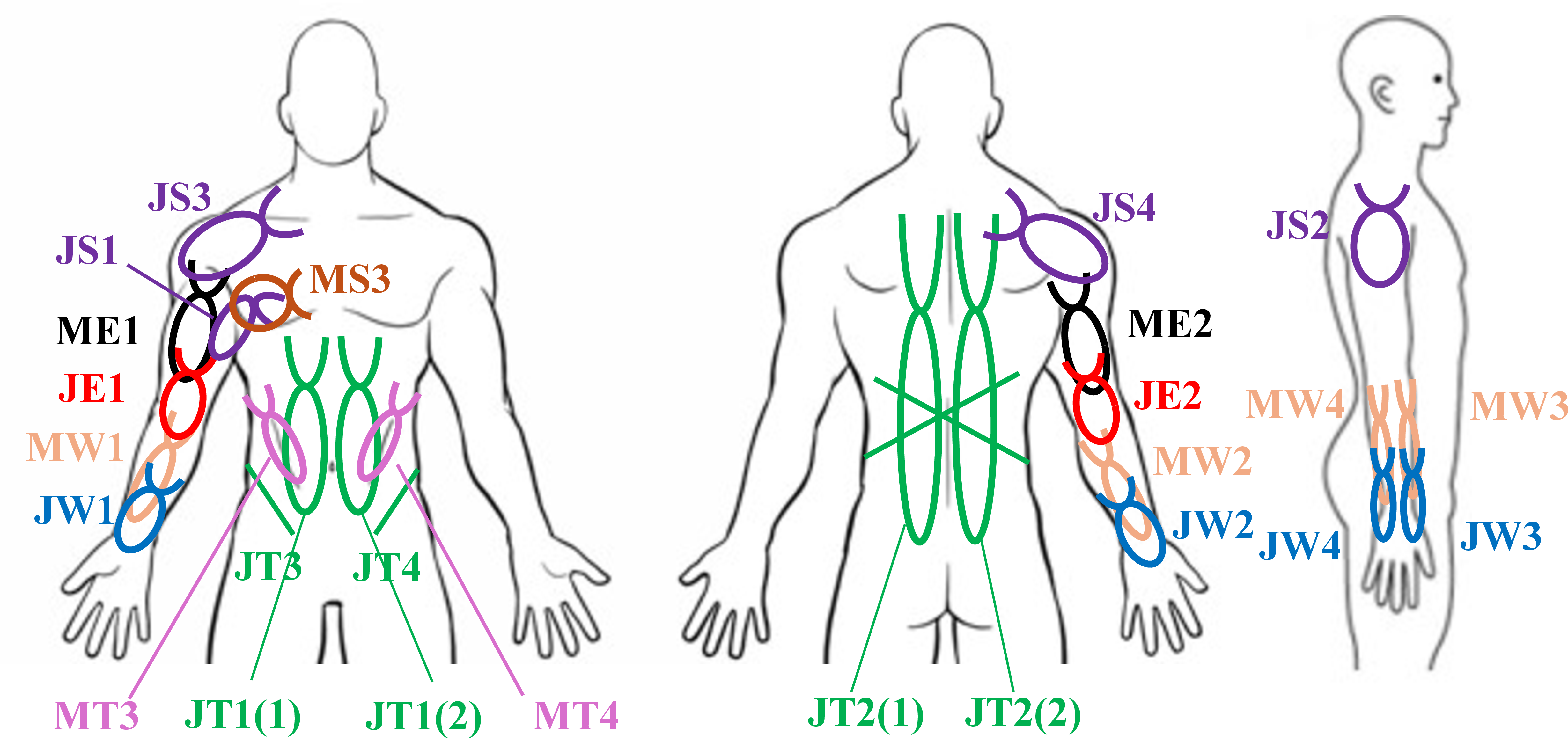}
    \caption{Configuration of artificial muscles and the definition of stimuli name. The artificial muscles of JT3 and JT4 (Trunk Right/Left Rotation) are linear shape and placed so that they spans the front and back of the trunk}
    \label{g3}
\end{figure*}

Funabot-Upper consists of 17 artificial muscles in each upper limb and 8 in the trunk, for a total of 42. 
Fig. \ref{g3} shows the 25 artificial muscle configurations corresponding to the right upper limb and trunk. The arrangement of the artificial muscle of the left upper limb is symmetric to the midline with that of the right upper limb. Artificial muscles are placed in an elliptical shape and can present the shear force to the center of the elliptical shape.
In a previous study, it was found that circular shape provided higher accuracy in perceiving motion direction compared to cross or parallel shape, and that shear force of circular shape is located at the starting point of the pressing force\cite{Yamaguchi-Robomech}. 
Thus, most artificial muscles are ellipse shape. Only for the artificial muscles arrangement corresponding to trunk left rotation and right rotation, a linear arrangement was adopted due to the limited arrangement area.

%--------------------------------------------------%

%--------------------------------------------------%

\begin{figure*}[!t]
    \centering
    \vspace{2mm}
    \includegraphics[width=0.7\linewidth]{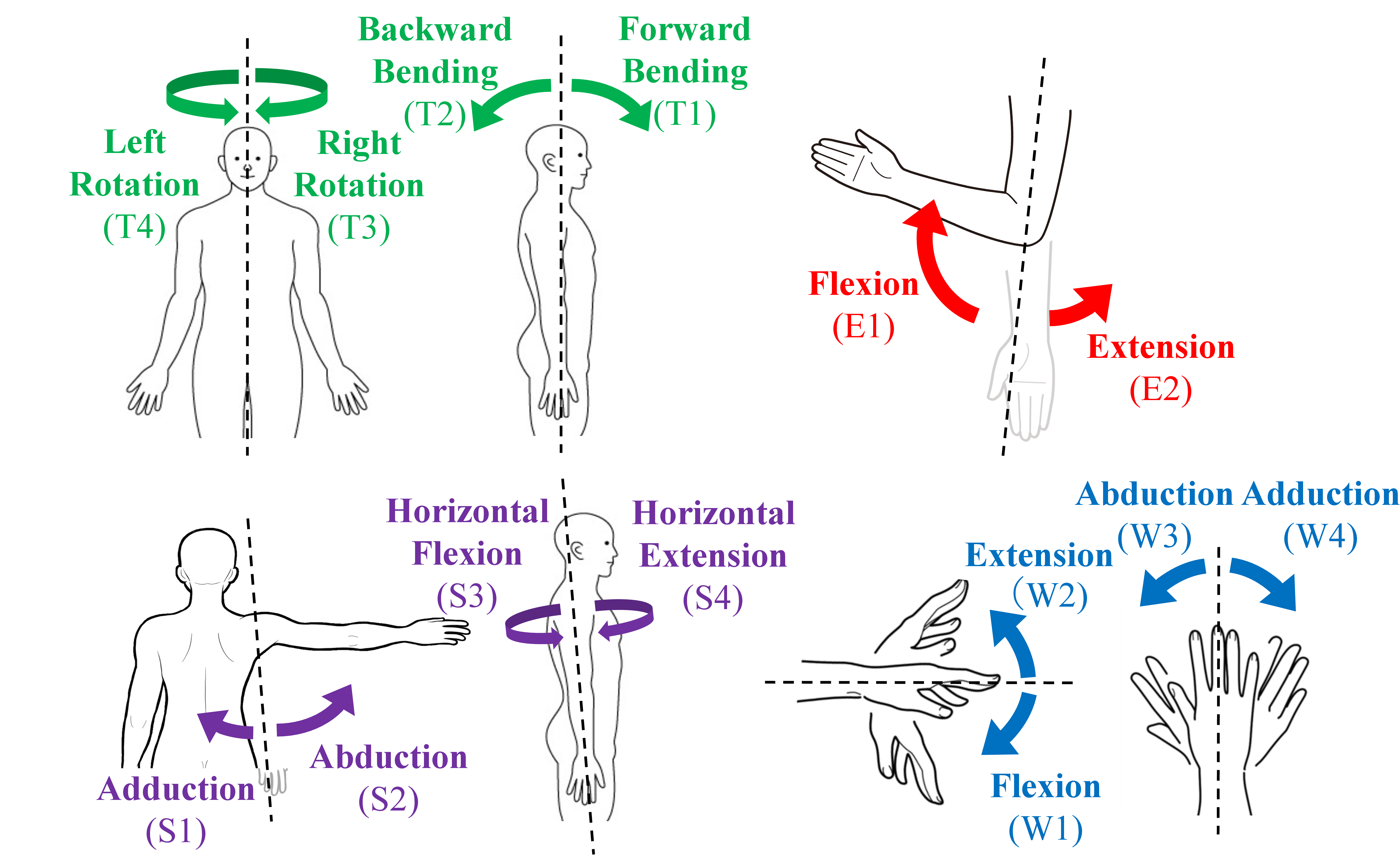}
    \caption{The definition of the target 14 motions. Each of the motions is assigned a name. The first letter (T, S, E, and W) represents the target body part (Trunk, Shoulder, Elbow, and Wrist)}
    \label{g2}
\end{figure*}

Each of the 23 stimuli shown in Fig. \ref{g2} is assigned a name according to its design policy and target motion. The first letter represents the design policy (J or M), and the second and third letters represent the target motion, as shown in Fig. \ref{g2}.
The target muscles of Type M stimuli are shown in Table \ref{t1}\cite{MT3and4,MS3,ME1,ME2,MW1,MW2,MW3,MW4}.

%---------------------%
%---------------------%

\subsection{The Fabrication of Funabot-Upper}
Funabot-Upper shown in Fig. \ref{g1}(a) is designed to induce the kinesthetic perception of 14 motions in the trunk, shoulder, elbow, and wrist based on our new design policy that mentioned at section II.A. The 14 target motions are shown in Fig. \ref{g2}. It consists of 42 thin McKibben artificial muscles (EM20, s-muscle) placed on polyester underwear (12JA5P1009, Mizuno Corporation) and gloves. The underwear has high elasticity and fits the wearer's body. Underwear and gloves are connected through artificial muscles.

%--------------------------------------------------------------%
\begin{table}
\centering
\caption{\scriptsize{MUSCLES OF UPPER BODY AND THE MOTIONS THEY CONTRIBUTE}}
\label{t1}
\begin{tabular}{ l l l }
\hline
Muscle of Upper Body & Contributing Motion & Pattern\\
\hline
External Oblique & Trunk Right Rotation & MT3\\
External Oblique & Trunk Left Rotation & MT4\\ 
Pectoralis Major & Shoulder Horizontal Flexion & MS3\\
Biceps Brachii & Elbow Flexion & ME1\\
Triceps Brachii & Elbow Extension & ME2\\
Palmaris longus & Wrist Flexion & MW1\\
Extensor Digitorum & Wrist Extension & MW2\\
Flexor Carpi Radialis & Wrist Abduction & MW3\\
Flexor Carpi Ulnaris & Wrist Adduction & MW4\\
\hline 
\end{tabular}
\end{table}
%-----------------------------------------------------------------%

Based on our previous study of activating fabric\cite{nakagawa2022turning}, artificial muscles are embroidered on the suits in two different ways shown in Fig. \ref{g1}(b). One type is highly constraining due to its narrow embroidery width and transfers the contraction force of the artificial muscles to the suit. The other type has a wider embroidery width and suppresses misalignment of the artificial muscles. Each artificial muscle placed on the Funabot-Upper is arranged in the ellipse muscle arrangement. The endpoints of the artificial muscle and the lower edge of the ellipse were made of narrow embroidery width type, while the rest of the muscle was made of the wide embroidery width type, with the expectation that the contraction force would be strongly transmitted in the long axis direction. Both embroidered regions are fixed by the expansion of the artificial muscle when pneumatic pressure is applied. Therefore, the length of the artificial muscles placed on the suit can be adjusted in the non-inflated state, allowing it to accommodate the physiques of various wearers. 

Fig. \ref{design}(d) shows the artificial muscle processed for use in Funabot-Upper. After inserting a total of four 4 mm polyurethane tubes (KDU4-CB-100, Koganei) cut to 1 cm width into the artificial muscle, insert a 1.8 mm polyurethane tube (U2-C-100, Koganei) into one side of the artificial muscle. Next, cover both ends of the artificial muscle with the 4 mm tubes at each end and seal the air escape route at the endpoints using instant glue. This process ensures air is delivered through the 1.8 mm tube without leakage. Furthermore, sliding the central 4 mm tube allows adjustment to the required length for the arrangement of each artificial muscle.

\subsection{Control System}
\begin{figure*}[!t]
    \centering
    \includegraphics[width=1\linewidth]{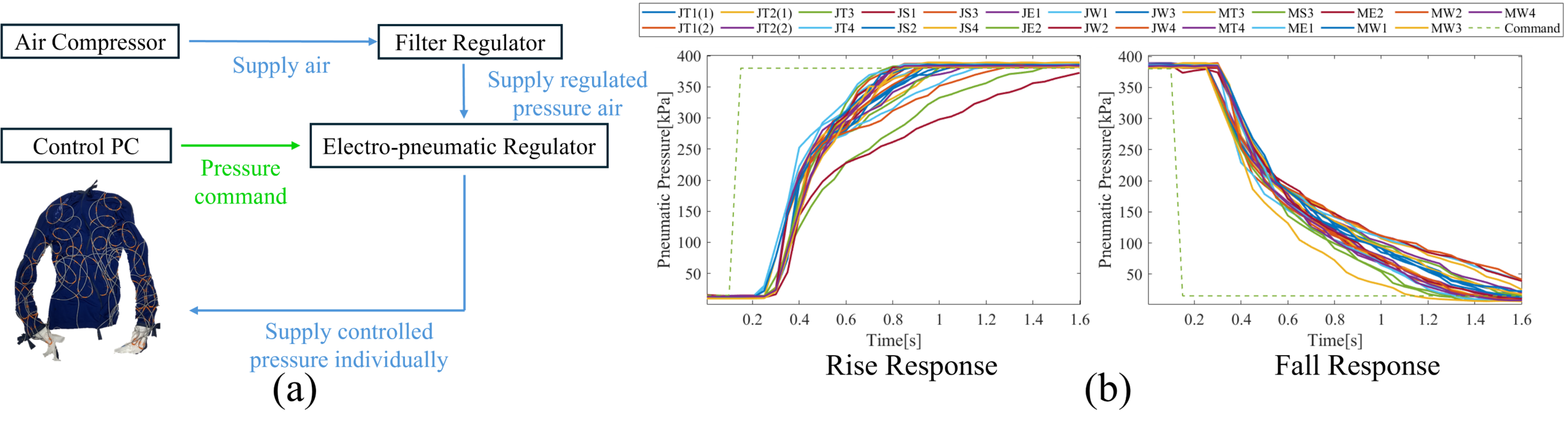}
    \caption{(a) Control system of suit. The filter regulator can rectify the air (blue line) from the air compressor to a specific pressure value before inducing it to the electro-pneumatic regulator. The PC can control the pneumatic pressure applied to the artificial muscles on the suit through the command line (orange line), thereby controlling the artificial muscles. (b) Time variation applied pneumatic pressure on artificial muscles.}
    \label{g4}
\end{figure*}

A schematic diagram of the system that controls the contraction of the artificial muscles on the suit is shown in Fig. \ref{g4}(a). The system consists of four components: an air compressor that supplies the pneumatic pressure needed to drive artificial muscles, a filter regulator that adjusts the pneumatic pressure and supply regulated pressure air, an electro-pneumatic regulator that controls the pneumatic pressure supplied to each artificial muscle, and a control PC that executes the logic of the system. The specifications of each module are provided in Table \ref{t3}. 

%-------------------------------------------%
\begin{table*}[!t]
\centering
\caption{SPECIFICATION OF THE MODULES ON THE CONTROL SYSTEM}
\label{t3}
\begin{tabular}{ l  c  c  c }
\hline
Parameters & Compressor & Filter Regulator & Electro-pneumatic Regulator\\
\hline
Vendor & PAOCK & Koganei & Koganei\\
Model Number & SOL-1030 & FRZB30 & CRCB-0135W/0136W\\
Maximum Pneumatic Pressure & 0.80 MPa & 1.0 MPa & -\\
Output Air Speed & 84 L/min & - & -\\
Resolution of Pneumatic Pressure & - & - & \(\pm\)0.1 kPa\\
\hline 
\end{tabular}
\end{table*}

%--------------------------------------------------%

The 42 artificial muscles used in Funabot-Upper are connected to the electro-pneumatic regulator one to one. The control program is developed in C++ using Microsoft Visual Studio. The logic controllers (M-DUINO PLC ETHERNET 54 I/Os Analog/Digital Plus from Industrial Shields) are installed to convert command pressure values from the control PC into analog voltage values for inputting to electro-pneumatic controller. This module receives the pneumatic command string from the control PC via wired LAN, converts it to an analog voltage, and outputs it to the regulators. The pressure values applied to muscles are controlled at 380 kPa and 15 kPa for the contraction and relaxation state, respectively.

The results of the rise response and the fall response of the system are shown in Fig. \ref{g4}(b) for the 25 artificial muscles corresponding to the right upper limb and trunk. The horizontal and vertical axes represent time and pressure values, respectively.
The time response of the pressure value to the command was obtained by measuring the output of the pressure sensor corresponding to the pressure value supplied to the artificial muscle.
The dotted lines labeled “Command” depict the pressure command inputted into the artificial muscles, while other lines represent the actual pressure values applied to the muscles.
The graphs show that the response lags behind the pressure command.
The rise response is 0.896 ± 0.212 s and the fall response is 1.442 ± 0.196 s.
The fall response is relatively slow because the muscles relax in accordance with the release of pressure into the atmosphere.
In the experiments of this study, we evaluate induced kinesthetic perception under conditions where stimuli are presented for a sufficient duration. Therefore, this delay is not an issue.

\section{EXPERIMENT 1: EVALUATION OF KINESTHETIC PERCEPTION BY SINGLE STIMULUS}
An experiment was conducted to evaluate the kinesthetic perception induced by a single stimulus. In the experiment, single stimulus placed on the joint and muscle areas was presented randomly in standing position, and subjects were asked to respond to the motion they perceive. The Ethics Committee of the Graduate School of Engineering, Nagoya University, approved the experiments (Approval Number: 24-10).

%--------------------------------------------------%
\begin{figure}[t]
    \centering
    \includegraphics[width=0.6\linewidth]{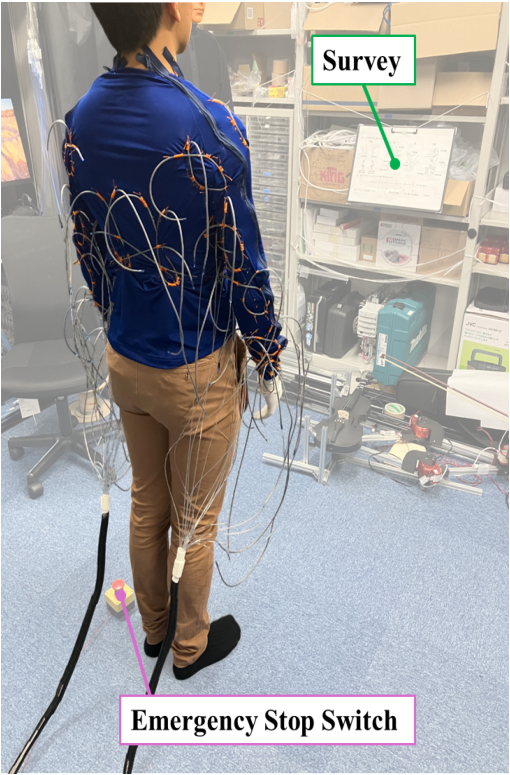}
    \caption{The appearance of the subject in the experiment. They look the survey in front of them and the emergency stop button is placed at their feet.}
    \label{g5}
\end{figure}
%--------------------------------------------------%

\subsection{Experimental Setup}
The experiment was conducted on eight healthy male participants in their 20s with relatively close body parameters (average height: 1.68 ± 0.03 m; weight: 58.4 ± 3.2 kg; BMI: 20.8 ± 1.1) to reduce the influence of size.

Fig. \ref{g5} shows the experimental environment. During each trial, a subject wore the Funabot-Upper suit while standing and was instructed to look at a questionnaire sheet positioned in front of them in order to prevent them from perceiving the behavior of the Funabot-Upper visually.

Each of the 23 designed stimuli shown in Fig. \ref{g3} (14 Type J and 9 Type M) was presented once per set in a random order, and each participant completed three sets (69 trials total).
Subjects reported one motion they perceived from the 14 options shown in Fig. \ref{g2}.
No training phase was conducted during the experiment.

\subsection{Result}
\begin{figure*}[!t]
    \centering
    \vspace{2mm}
    \includegraphics[width=1\linewidth]{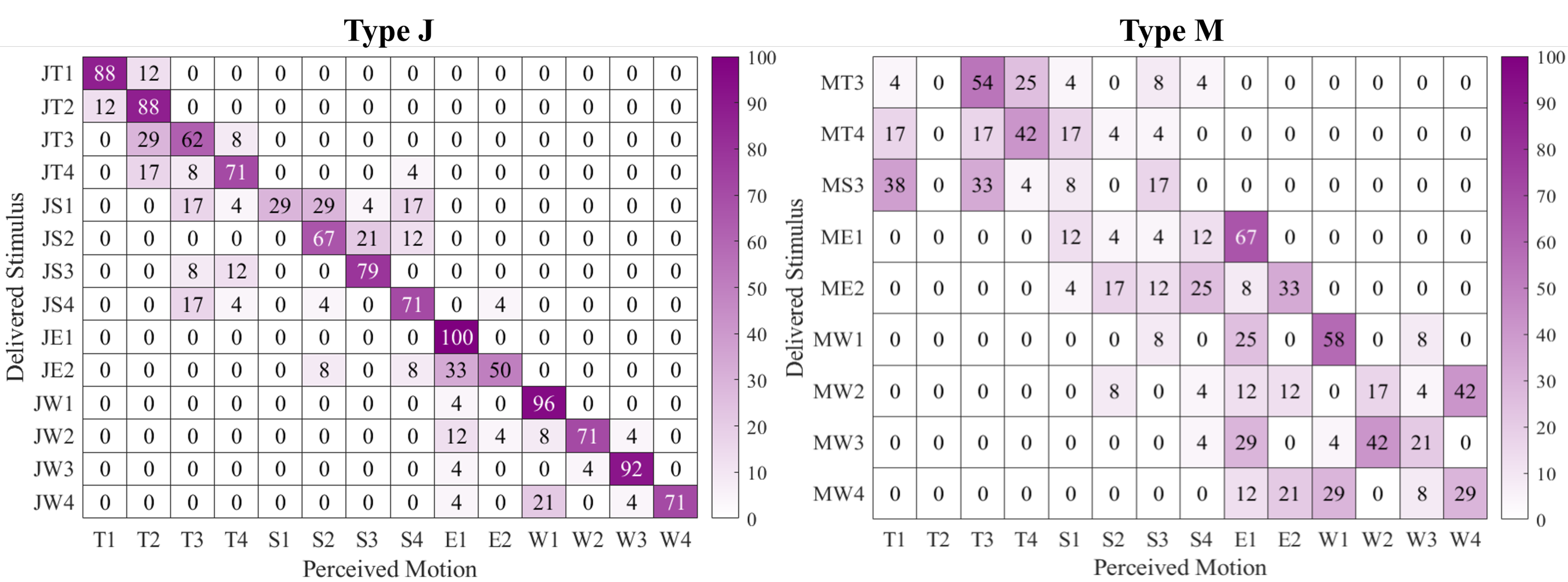}
    \caption{Confusion matrices in the Experiment 1. User response percentages are shown in these confusion matrices. Type J represents the stimulus delivered to the joint, and Type M represents the stimulus delivered to the muscle. The vertical axis indicates the delivered stimulus and the horizontal axis indicates the perceived motion by subjects; the higher the percentage, the darker the purple shade used to fill that cell.}
    \label{g6}
\end{figure*}

Fig. \ref{g6} shows the confusion matrices of perceived motion. The vertical axis represents the type of stimulus presented to the subject, and the horizontal axis represents the motion actually perceived by the subject. 

%--------------------------------------------------%
%--------------------------------------------------%

For Type J stimuli, the motion with the highest perception rate was elbow flexion (E1) with a perception rate of 100\%, whereas shoulder adduction (S1) and elbow extension (E2) were perceived at rates of 50\% or less.
The average correct recognition rate was 73.8\%, with 77.3\% for the trunk, 61.5\% for the shoulder, 75.0\% for the elbow, and 82.5\% for the wrist motions. 
The percentage of perception as the intended motion was higher than the coincidence rate (7.14\%) for all stimuli and the coincidence rate of each region (trunk: 25\%, shoulder: 25\%, elbow: 50\%, wrist: 25\%). 
When considering only body-part identification regardless of direction, the accuracy reached 90.7\%.

For Type M stimuli, the motion with the highest perception rate was elbow flexion (E1) with a perception rate of 67\%, while shoulder horizontal flexion (S3) and wrist extension (W2) showed the worst results of 17\%.
The average percentages perceived as intended motion were 37.5\%. The percentage of perception as the intended motion was higher than the coincidence rate (7.14\%) for all stimuli, but was lower than that of Type J.
When focusing on which body part was perceived as the intended motion without distinguishing the motion direction, the body part was correctly perceived 60.6\%.

\subsection{Discussion}
Type J stimuli can be perceived at relatively high rate of about 70\%. In addition, when focusing on the part of the body that perceives motion in Type J stimuli, subjects perceive about 90\%. This shows that Type J stimuli can be perceived without mixing with motion in other body parts. 
In the experiment, subjects maintained a standing position with their arms extended straight down. Therefore, motion of E2 and S1 could not be moved any further, which likely resulted in low perception rates.

As for Type M, it can be perceived at about 40\%. It supports the hypothesis that the perceptual mixing of Funabot-Suit for Upper Body can be caused by stimulating muscle. It is considered that perceptual mixing of Funabot-Suit for Upper Body resulted from unclear stimuli provided in multiple body parts simultaneously. 
On the other hand, the perceptual tendencies suggest that Type M stimuli can be made to perceive motion by presenting the sensation of the stimulated area being pulled. 
For example, MW1, placed on the palm side of the forearm, was perceived palm side of the forearm being pulled as E1 (elbow flexion).  This tendency is also common to MS3–T1, ME1–S1, ME2–S2, ME2–S4, MW2–E2, MW4–E1 (Delivered Stimulus-Perceived Stimulus). It suggests that the two opposing shear forces present ``pinching" sensation\cite{Funabot2} and may induce a feeling of being pulled in that area.

\section{EXPERIMENT 2: COMPARISON EXPERIMENT WITH PREVIOUS STUDY}
 In Experiment 1, we confirmed the performance of Funabot-Upper. In this experiment, we compared performance between Funabot-Upper and Funabot-Suit for Upper Body. This experiment was conducted under the ethical review (approval number: 24-10) as in Experiment 1. Evaluation was made by a survey in which subjects were asked to describe the motion they perceived.

%--------------------------------------------------%
\begin{figure*}[!t]
    \centering
    \vspace{2mm}
    \includegraphics[width=1\linewidth]{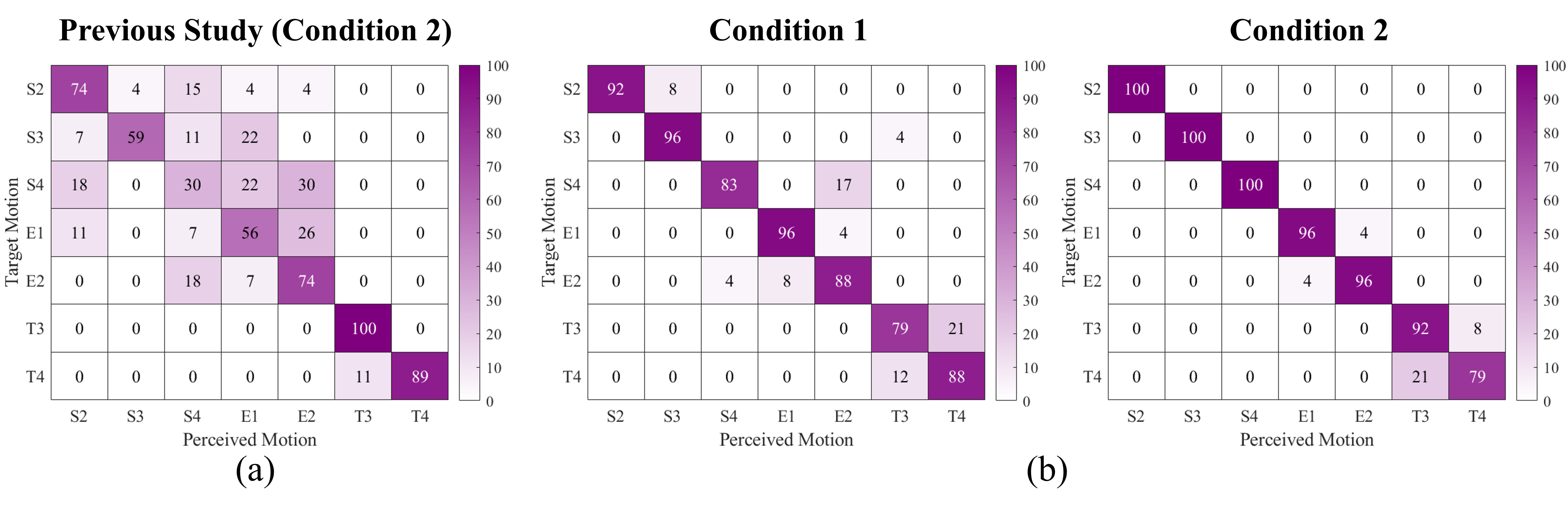}
    \caption{(a)Confusion matrix in the experiment of previous study (Funabot-Suit for Upper Body).  (b)Confusion matrices in the Experiment 2. User response percentages are shown in these confusion matrices. The vertical axis indicates the target motion (also see Table \ref{t5} for stimuli used in Experiment 2) and the horizontal axis indicates the perceived motion by subjects; the higher the percentage, the darker the purple shade used to fill that cell.}
    \label{g9}
\end{figure*}
%--------------------------------------------------%

\subsection{Experimental Setup}
This experiment was conducted after Experiment 1. The subjects were same members as Experiment 1. When the stimuli were presented, subjects were requested to select the perceived motion from 7 motions. To conduct the experiment under the same conditions as previous studies, unlike Experiment 1, stimulation was applied simultaneously to both sides of the artificial muscle to generate the target motion. The experimental environment is the same as Experiment 1.

Based on Experiment 1, we hypothesized that it can induce more intuitive kinesthetic perception by combining stimulation of the muscles and joints. While there are multiple ways to combine stimulation, we tried combining stimulation using the ``pinching" sensation made by Type M stimuli.  
The target 7 motions and the stimuli used in this experiment were listed in Table \ref{t5}. This experiment was conducted under two conditions. The first is the condition that subjects are not given any prior information (Condition 1). The second is the condition that subjects are given correct answer of stimuli (Condition 2). Condition 2 is the same as that used in the experiment of the previous study. The experiment of Condition 2 was conducted after that of Condition 1. 

Each of stimuli was presented one time at random in a set. Each subject performed three sets of trials (there were 21 trials for each condition). 

\subsection{Result and Discussion}
Fig. \ref{g9} shows the experimental results. 
Under Condition 1 (without prior feedback), subjects achieved 88.7\% accuracy, and under Condition 2 (with correct feedback), accuracy increased to 94.6\%. Compared to the 68.8\% achieved with the previous Funabot-Suit, this represents a substantial improvement in perception accuracy.
Significant improvements in perception rate were observed in the shoulder and elbow regions where densely distributed muscles often cause perceptual mixing.
These improvements can be attributed to the new design policy, which separates joint and muscle stimulation to reduce interference.
Type J stimuli and limiting stimulation area may have made it easier for the wearer to distinguish between specific stimuli corresponding to the motion direction and stimuli that trigger other motion directions.
As a result, Funabot-Upper can induce more accurate kinesthetic perception without training.  

%--------------------------------------------------%
\begin{table}
\centering
\caption{Combinations of stimuli used in Experiment 2}
\label{t5}
\begin{tabular}{ l l l }
\hline
Name & Target Motion & Stimuli\\
\hline
\hline
S2 & Shoulder Abduction & JS2\\
S3 & Shoulder Horizontal Flexion & JS3+ME1\\
S4 & Shoulder Horizontal Extension & JS4+ME2\\
E1 & Elbow Flexion & JE1+MW1\\
E2 & Elbow Extension & JE2+MW2\\
T3 & Trunk Right Rotation & JT3+MS3\\
T4 & Trunk Left Rotation & JT4+MS3\\

\hline 
\end{tabular}
\end{table}
%--------------------------------------------------%

The results of S4 and E2 performed under Condition 1 were lower than those of other shoulder or elbow motions. The reasons for the E2 results are thought to be the same as for Experiment 1: restricted range of motion due to the standing position. The S4 results are thought to be because of motion direction. While S4 originally involved motion in the same direction as E2, combining it with ME2 stimulation caused the entire arm to move backward, making it difficult to distinguish from the E2 motion.
By giving the correct answer, these motions could be distinguished. 
The perception rate of trunk motion decreases compared to the previous study.
It may be due to a decrease in the number of artificial muscles stimulating trunk. Funabot-Upper used one artificial muscle for each trunk right rotation and trunk left rotation, while Funabot-Suit for Upper Body used 10 for each motion.
The perception rate of trunk may be improved by increasing the number of artificial muscles.

In this experiment, 26 of 42 artificial muscles were used to induce kinesthetic perception of 7 motions, while previous study used 80 artificial muscles. We achieved a 67.5\% reduction in the number of artificial muscles while improving the perception rate. This leads to system simplification, cutting production cost, and lightening suit weight.

\section{Conclusion}
This study introduced Funabot-Upper, a McKibben muscle–actuated haptic suit capable of presenting 14 distinct upper-body motions in the trunk, shoulder, elbow, and wrist.
By arranging elliptical shaped artificial muscles on clothing, the suit can stimulate human joints and muscles independently.
Experiments revealed that joint stimulation produces distinct kinesthetic sensations without perceptual mixing, while muscle stimulation tends to evoke a pulling sensation in the stimulated area.
By combining joint and muscle stimuli, the proposed system reduced perceptual interference and improved recognition accuracy from 68.8\% to 94.6\% compared with the previous Funabot-Suit. 

Future work will focus on optimizing the response time of pneumatic actuation, improving perception rates for muscle stimuli, and exploring combined control strategies for more natural kinesthetic feedback. These results provide a foundation for developing practical, full-body wearable haptic systems for immersive and assistive applications.

\bibliographystyle{IEEEtran}
\bibliography{ref}

@article{Wrist1_Vibe1,
  title={Degree of muscle-and-tendon tonus effects on kinesthetic illusion in wrist joints toward advanced rehabilitation robotics},
  author={Komura, Hiraku and Kubo, Takumu and Honda, Masakazu and Ohka, Masahiro},
  journal={Robotica},
  volume={40},
  number={4},
  pages={1222--1232},
  year={2022},
  publisher={Cambridge University Press}
}

@article{Wrist2,
  title={A wearable wrist haptic display for motion tracking and force feedback in the operational space},
  author={Laghi, Marco and Catalano, Manuel G and Grioli, Giorgio and Bicchi, Antonio},
  journal={Wearable Technologies},
  volume={2},
  pages={e5},
  year={2021},
  publisher={Cambridge University Press}
}

@inproceedings{Elbow1,
  title={Simultaneous control of tonic vibration reflex and kinesthetic illusion for elbow joint motion toward novel robotic rehabilitation},
  author={Kiguchi, Kazuo and Maemura, Kanta},
  booktitle={2021 43rd Annual International Conference of the IEEE Engineering in Medicine \& Biology Society (EMBC)},
  pages={4773--4776},
  year={2021},
  organization={IEEE}
}

@article{Elbow2,
  title={Smart textiles that teach: fabric-based haptic device improves the rate of motor learning},
  author={Ramachandran, Vivek and Schilling, Fabian and Wu, Amy R and Floreano, Dario},
  journal={Advanced Intelligent Systems},
  volume={3},
  number={11},
  pages={2100043},
  year={2021},
  publisher={Wiley Online Library}
}

@inproceedings{Shoulder1,
  title={Performance analysis of vibrotactile and slide-and-squeeze haptic feedback devices for limbs postural adjustment},
  author={Lorenzini, Marta and Ciotti, Simone and Gandarias, Juan M and Fani, Simone and Bianchi, Matteo and Ajoudani, Arash},
  booktitle={2022 31st IEEE International Conference on Robot and Human Interactive Communication (RO-MAN)},
  pages={707--713},
  year={2022},
  organization={IEEE}
}

@inproceedings{Shoulder2,
  title={A study on the generation of kinesthetic illusion, tonic vibration reflex, and antagonist vibratory response in the shoulder joint extension direction by vibration stimulation to the origin and insertion in the biceps brachii muscle},
  author={Teranishi, Taiki and Nishikawa, Satoshi and Kiguchi, Kazuo},
  booktitle={2023 IEEE/SICE International Symposium on System Integration (SII)},
  pages={1--6},
  year={2023},
  organization={IEEE}
}

@article{Trunk1,
  title={Vibrotactile feedback to make real walking in virtual reality more accessible},
  author={Mahmud, M Rasel and Stewart, Michael and Cordova, Alberto and Quarles, John},
  journal={arXiv preprint arXiv:2208.02403},
  year={2022}
}

@article{Trunk2,
  title={Evaluation of presence in virtual environments: haptic vest and user’s haptic skills},
  author={Garc{\'\i}a-Valle, Gonzalo and Ferre, Manuel and Bre{\~n}osa, Jose and Vargas, David},
  journal={IEEE Access},
  volume={6},
  pages={7224--7233},
  year={2017},
  publisher={IEEE}
}

@inproceedings{Vibe2,
  title={Simultaneous control of tonic vibration reflex and kinesthetic illusion for elbow joint motion toward novel robotic rehabilitation},
  author={Kiguchi, Kazuo and Maemura, Kanta},
  booktitle={2021 43rd Annual International Conference of the IEEE Engineering in Medicine \& Biology Society (EMBC)},
  pages={4773--4776},
  year={2021},
  organization={IEEE}
}

@inproceedings{Vibe3,
  title={A study on the generation of kinesthetic illusion, tonic vibration reflex, and antagonist vibratory response in the shoulder joint extension direction by vibration stimulation to the origin and insertion in the biceps brachii muscle},
  author={Teranishi, Taiki and Nishikawa, Satoshi and Kiguchi, Kazuo},
  booktitle={2023 IEEE/SICE International Symposium on System Integration (SII)},
  pages={1--6},
  year={2023},
  organization={IEEE}
}

@article{Vibe4,
  title={The effects of periodic and noisy tendon vibration on a kinesthetic targeting task},
  author={Eschelmuller, Gregg and Szarka, Annika and Gandossi, Braelyn and Inglis, J Timothy and Chua, Romeo},
  journal={Experimental Brain Research},
  volume={242},
  number={1},
  pages={59--66},
  year={2024},
  publisher={Springer}
}

@article{Skin1,
  title={Design and evaluation of a wearable skin stretch device for haptic guidance},
  author={Chinello, Francesco and Pacchierotti, Claudio and Bimbo, Joao and Tsagarakis, Nikos G and Prattichizzo, Domenico},
  journal={IEEE Robotics and Automation Letters},
  volume={3},
  number={1},
  pages={524--531},
  year={2017},
  publisher={IEEE}
}

@article{Skin2,
  title={Rotational motion due to skin shear deformation at wrist and elbow},
  author={Nakamura, Takuto and Kuzuoka, Hideaki},
  journal={IEEE Transactions on Haptics},
  volume={17},
  number={1},
  pages={108--115},
  year={2024},
  publisher={IEEE}
}

@inproceedings{Skin3,
  title={A wearable skin stretch device for haptic feedback},
  author={Bark, Karlin and Wheeler, Jason and Lee, Gayle and Savall, Joan and Cutkosky, Mark},
  booktitle={World Haptics 2009-Third Joint EuroHaptics conference and Symposium on Haptic Interfaces for Virtual Environment and Teleoperator Systems},
  pages={464--469},
  year={2009},
  organization={IEEE}
}

@article{Skin4,
  title={The cuff, clenching upper-limb force feedback wearable device: Design, characterization and validation},
  author={Barontini, Federica and Catalano, Manuel G and Fani, Simone and Grioli, Giorgio and Bianchi, Matteo and Bicchi, Antonio},
  journal={IEEE Transactions on Haptics},
  volume={17},
  number={4},
  pages={662--675},
  year={2024},
  publisher={IEEE}
}

@article{nakagawa2022turning,
  title={Turning a functional cloth into an actuator by combining thread-like thin artificial muscles and embroidery techniques},
  author={Nakagawa, Koki and Sakai, Yusuke and Funabora, Yuki and Doki, Shinji},
  journal={IEEE Robotics and Automation Letters},
  volume={7},
  number={3},
  pages={5827--5833},
  year={2022},
  publisher={IEEE}
}

@inproceedings{Funabot1,
  title={Flexible fabric actuator realizing 3D movements like human body surface for wearable devices},
  author={Funabora, Yuki},
  booktitle={2018 IEEE/RSJ International Conference on Intelligent Robots and Systems (IROS)},
  pages={6992--6997},
  year={2018},
  organization={IEEE}
}

@inproceedings{Funabot2,
  title={Verification of Mehcanism How Fabric Actuator Evoke Pinched Tactile by Using Force Distribution Sensor},
  author={Masaoka, Shinichi and Sato, Yusei and Funabora, Yuki and Doki, Shinji},
  booktitle={2024 IEEE International Conference on Cyborg and Bionic Systems (CBS)},
  pages={263--266},
  year={2024},
  organization={IEEE}
}

@article{MT3and4,
  title={Functional roles of abdominal and back muscles during isometric axial rotation of the trunk},
  author={Ng, Joseph K-F and Parnianpour, Mohamad and Richardson, Carolyn A and Kippers, Vaughan},
  journal={Journal of Orthopaedic Research},
  volume={19},
  number={3},
  pages={463--471},
  year={2001},
  publisher={Wiley Online Library}
}

@article{ME1,
  title={Electrophysiological Studies of the Biceps Brachii Activities in Supination and Flexion of the Elbow Joint},
  author={AKIRA NAITO and MICHIHIRO YAJIMA and HIDEHIKO FUKAMACHI and KOJI USHIKOSHI and YASUNOBU HANDA and NOZOMU HOSHIMIYA and YOSHIFUSA SHIMIZU},
  journal={The Tohoku Journal of Experimental Medicine},
  volume={173},
  number={2},
  pages={259-267},
  year={1994},
  doi={10.1620/tjem.173.259}
}

@article{ME2,
  title={The different role of each head of the triceps brachii muscle in elbow extension},
  author={Kholinne, Erica and Zulkarnain, Rizki Fajar and Sun, Yu Cheng and Lim, SungJoon and Chun, Jae-Myeung and Jeon, In-Ho},
  journal={Acta orthopaedica et traumatologica turcica},
  volume={52},
  number={3},
  pages={201--205},
  year={2018},
  publisher={Elsevier}
}

@article{MS3,
  title={An electromyographic analysis of functional differentiation in human pectoralis major muscle},
  author={Paton, ME and Brown, JMM},
  journal={Journal of Electromyography and Kinesiology},
  volume={4},
  number={3},
  pages={161--169},
  year={1994},
  publisher={Elsevier}
}

@article{MW1,
  title={Revisiting the functional anatomy of the palmaris longus as a thenar synergist},
  author={Moore, Colin W and Fanous, Jacob and Rice, Charles L},
  journal={Clinical Anatomy},
  volume={31},
  number={6},
  pages={760--770},
  year={2018},
  publisher={Wiley Online Library}
}

@article{MW2,
  title={Characterizing forearm muscle activity in young adults during dynamic wrist flexion--extension movement using a wrist robot},
  author={Forman, Davis A and Forman, Garrick N and Avila-Mireles, Edwin J and Mugnosso, Maddalena and Zenzeri, Jacopo and Murphy, Bernadette and Holmes, Michael WR},
  journal={Journal of Biomechanics},
  volume={108},
  pages={109908},
  year={2020},
  publisher={Elsevier}
}

@article{MW3,
  title={Characterizing forearm muscle activity in university-aged males during dynamic radial-ulnar deviation of the wrist using a wrist robot},
  author={Forman, Davis A and Forman, Garrick N and Avila-Mireles, Edwin J and Mugnosso, Maddalena and Zenzeri, Jacopo and Murphy, Bernadette and Holmes, Michael WR},
  journal={Journal of Biomechanics},
  volume={108},
  pages={109897},
  year={2020},
  publisher={Elsevier}
}

@article{MW4,
  title={Effects of forearm rotation on wrist flexor and extensor muscle activities},
  author={Ikeda, Kazuhiro and Kaneoka, Koji and Matsunaga, Naoto and Ikumi, Akira and Yamazaki, Masashi and Yoshii, Yuichi},
  journal={Journal of Orthopaedic Surgery and Research},
  volume={20},
  number={1},
  pages={53},
  year={2025},
  publisher={Springer}
}

@inproceedings{Yamaguchi-Robomech,
  title={Analysis of the Relationship Between Shoulder Motion Perception and Artificial Muscle Placement Patterns in Active Wear for Motion Instruction Based on Force Distribution},
  author={Yamaguchi, Takuma and Fukatsu, Haru and Funabora, Yuki and Doki, Shinji},
  booktitle={The Robotics and Mechatronics Conference},
  year={2025. (in Japanese)},
}

@article{Suit,
  title={Funabot-Suit: A bio-inspired and McKibben muscle-actuated suit for natural kinesthetic perception},
  author={Peng, Yanhong and Sakai, Yusuke and Nakagawa, Koki and Funabora, Yuki and Aoyama, Tadayoshi and Yokoe, Kenta and Doki, Shinji},
  journal={Biomimetic Intelligence and Robotics},
  volume={3},
  number={4},
  pages={100127},
  year={2023},
  publisher={Elsevier}
}

@article{yokoe2025intuitive,
  title={Intuitive Hand Positional Guidance Using McKibben-Based Surface Tactile Sensations to Shoulder and Elbow},
  author={Yokoe, Kenta and Funabora, Yuki and Aoyama, Tadayoshi},
  journal={IEEE Robotics and Automation Letters},
  year={2025},
  publisher={IEEE}
}

@inproceedings{yokoe2024elbow,
  title={Elbow angle guidance system based on surface haptic sensations elicited by lightweight wearable fabric actuator},
  author={Yokoe, Kenta and Aoyama, Tadayoshi and Funabora, Yuki and Takeuchi, Masaru and Hasegawa, Yasuhisa},
  booktitle={2024 IEEE International Conference on Advanced Intelligent Mechatronics (AIM)},
  pages={1447--1454},
  year={2024},
  organization={IEEE}
}

@article{electrical1,
  title={Haptic interface using tendon electrical stimulation with consideration of multimodal presentation},
  author={Takahashi, Akifumi and Tanabe, Kenta and Kajimoto, Hiroyuki},
  journal={Virtual Reality \& Intelligent Hardware},
  volume={1},
  number={2},
  pages={163--175},
  year={2019},
  publisher={Elsevier}
}

@inproceedings{Upper,
  title={Funabot-Suit for Upper Body: McKibben Actuated-Suit Inducing Seven Kinesthetic Perceptions in Elbow, Shoulder, and Trunk},
  author={Fukatsu, Haru and Peng, Yanhong and Funabora, Yuki and Doki, Shinji},
  booktitle={2025 IEEE International Conference on Systems, Man, and Cybernetics (SMC)},
  year={2025},
  organization={IEEE}
}

@article{rangwani2021new,
  title={A new approach of inducing proprioceptive illusion by transcutaneous electrical stimulation},
  author={Rangwani, Rohit and Park, Hangue},
  journal={Journal of NeuroEngineering and Rehabilitation},
  volume={18},
  number={1},
  pages={73},
  year={2021},
  publisher={Springer}
}

\end{document}